\documentclass[letterpaper, 10 pt, conference]{ieeeconf}
\IEEEoverridecommandlockouts      

\let\labelindent\relax

\usepackage{float}
\usepackage[utf8]{inputenc} 
\usepackage[T1]{fontenc}    
\usepackage{hyperref}       
\usepackage{url}            
\usepackage{booktabs}       
\usepackage{amsfonts}       
\usepackage{nicefrac}       
\usepackage{microtype}      
\usepackage{xcolor}         
\usepackage{xspace}
\usepackage{blindtext}
\usepackage{paralist}
\usepackage{tikz}
\usepackage{soul}
\usepackage{mathtools}
\usepackage{multirow}
\usepackage[inline]{enumitem}
\usepackage{algpseudocode}
\usepackage{comment}

\usepackage{algorithm}
\usepackage{amsmath} 
\usepackage{amssymb}  
\usepackage{graphicx}
\usepackage{caption}
\usepackage{subcaption}
\usepackage{booktabs}
\usepackage{longtable}

\makeatletter
\let\NAT@parse\undefined
\makeatother
\usepackage[numbers,compress]{natbib}

\DeclareMathOperator*{\argmax}{argmax}

\usepackage[capitalise]{cleveref}

\usetikzlibrary{calc}
\usetikzlibrary{fit}
\usetikzlibrary{positioning}
\usetikzlibrary{decorations.markings}
\usetikzlibrary{shapes.geometric}
\usetikzlibrary{shapes.multipart}
\usetikzlibrary{bayesnet}
\usetikzlibrary{tikzmark}
\usetikzlibrary{shapes}
\usetikzlibrary{3d}
\usetikzlibrary{arrows.meta}
\usetikzlibrary{shapes.geometric}

\definecolor{Red}{rgb}{1,0.05,0.05}
\definecolor{Green}{rgb}{0,0.8,0}
\definecolor{Blue}{rgb}{0,0,1}
\definecolor{LightBlue}{rgb}{0,0.5,1}
\definecolor{LightRed}{rgb}{1,0.25,0.25}
\definecolor{VeryLightRed}{rgb}{1,0.4,0.4}
\definecolor{ExtremelyLightRed}{rgb}{1,0.6,0.6}
\definecolor{Skin}{rgb}{1,0.71,0.69}
\definecolor{Grey}{rgb}{0.8,0.8,0.8}
\definecolor{LightGrey}{rgb}{0.6,0.6,0.6}
\definecolor{Black}{rgb}{0,0,0}
\definecolor{White}{rgb}{1,1,1}
\definecolor{MitRed}{rgb}{0.6,0.2,0.2}
\newcommand{\red}{\color{Red}}

\newcommand{\blue}{\color{Blue}}

\definecolor{green}{rgb}{0,0.69,0}
\definecolor{brightGreen}{rgb}{0,0.9,0}
\definecolor{pink}{RGB}{235,0,235}
\definecolor{blue}{RGB}{0,80,255}
\definecolor{brightBlue}{RGB}{0,0,255}
\definecolor{slateblue}{RGB}{71,60,139}
\definecolor{dullyellow}{RGB}{145,145,0}

\usepackage{fancyhdr}

\fancypagestyle{TitlePage}{
\lfoot{\emph{Proceedings of the 39th International Conference on Robotics and Automation (ICRA 2022).}}
}

\title{Incorporating Rich Social Interactions Into MDPs}

\author{Ravi Tejwani$^{1*}$, Yen-Ling Kuo$^{1*}$, Tianmin Shu$^{2}$, Bennett Stankovits$^{1}$\\Dan Gutfreund$^{3}$, Joshua B. Tenenbaum$^{2}$, Boris Katz$^{1}$, Andrei Barbu$^{1}$
  \thanks{$^{*}$ Equal contribution}
  \thanks{$^{1}$ CSAIL \& CBMM, MIT}
  \thanks{\hspace*{2ex}\texttt{\{tejwanir,ylkuo,bstankov,boris,abarbu\}@mit.edu}}
  \thanks{$^{2}$ BCS \& CBMM, MIT \texttt{\{tshu,jbt\}@mit.edu}}
  \thanks{$^{3}$ MIT-IBM Watson AI Lab \texttt{dgutfre@us.ibm.com}}%
}

\begin{document}
\maketitle
\thispagestyle{TitlePage}

\begin{abstract}
  Much of what we do as humans is engage socially with other agents, a skill
  that robots must also eventually possess.
  We demonstrate that a rich theory of social interactions originating from
  microsociology and economics can be formalized by extending a nested MDP where
  agents reason about arbitrary functions of each other's hidden rewards.
  This extended Social MDP allows us to encode the five basic interactions that
  underlie microsociology: cooperation, conflict, coercion, competition, and
  exchange.
  The result is a robotic agent capable of executing social interactions
  zero-shot in new environments; like humans it can engage socially in novel
  ways even without a single example of that social interaction.
  Moreover, the judgments of these Social MDPs align closely with those of
  humans when considering which social interaction is taking place in an
  environment.
  This method both sheds light on the nature of social interactions, by
  providing concrete mathematical definitions, and brings rich social
  interactions into a mathematical framework that has proven to be natural for
  robotics, MDPs.
\end{abstract}

\section{Introduction}

Endowing robots with the ability to understand and engage in social interactions
is key to having them integrate into our daily lives.
Yet, very little is known about social interactions, what they are, how to
measure them, and how to computationally implement them.
Research in psychology has attempted to define different types of social
interactions and to test the abilities of humans to engage in them \citep{goffman1978presentation, doi:10.1146/annurev.economics.050708.143312,  weber_1978, goffman1955face}.
While no single framework for what social interactions are has emerged out of
this work, one coherent proposal comes out of the field of microsociology; the
study of fine-grained face-to-face interactions.
Microsociology puts forward that underlying social interactions are five
fundamental abilities that are recombined to give rise to the repertoire of
behaviors we see in humans and other animals: cooperation, conflict,
competition, coercion, and exchange.
We provide the first computational mechanism for implementing these five
interactions.
Moreover, we put forward the first clear mathematical definition of what these
fives types of interactions are.
The result is a principled computational mechanism for social interactions that
allows robots to execute such interactions zero-shot in novel environments.

To implement these interactions, we build on top of \citet{tejwani2021social}, which recently
introduced Social MDPs --- an attempt at extending the MDP framework to social
interactions.
That work builds on an analogy, that the understanding of partial observability
took a great leap forward with the advent of POMDPs, and that perhaps doing the
same for social interactions by building Social MDPs will similarly pay off.
As executing MDPs is fairly efficient, this also ensures that the resulting
models are tractable.
Unfortunately, as originally formulated, Social MDPs are fundamentally limited
to two of the five interactions: cooperation (helping) and conflict (hindering).
Our new model is an extended Social MDP capable of handling these five
fundamental social interactions.
One feature of this model is that it degenerates to the original Social MDP
model for the two interactions that the two share, and furthermore, it
degenerates to a well-known social model in game theory for those two
interactions \citep{levine1998modeling}.

We make four contributions:
\begin{enumerate}
\item A novel Social MDP that allows robots to execute five fundamental social
  interactions from microsociology.
\item The first formalization of what those interactions are.
\item A computational implementation that enables robots to engage in social
  interactions zero-shot, without any social-interaction-specific training.
\item Extensive human experiments that validate the model and demonstrate its
  ability to capture human judgments about social interactions.
\end{enumerate}

\section{Related Work}

In multi-agent settings, an agent learns to reason about the goals, preferences
and beliefs of the other agents so that it can effectively interact with them
\citep{albrecht2018autonomous,he2016opponent}.
Several types of models have been explored in this space: theory-of-mind-based models for goal inference~\cite{baker2008theory,baker2014modeling,kiley2013mentalistic,kleiman2016coordinate,rabinowitz2018machine}, Bayesian inverse planning~\cite{baker2009action,ullman2009help}, and learning the reward functions of other agents~\cite{hadfield2016cooperative}.
\citet{xie2020learning} provide a method for learning a low-dimensional
representation of the strategy of another agent. This representation enables
agents to avoid or work with one another.
These methods have all been limited to social interactions exhibiting cooperation or conflict (helping or hindering).
Additionally, the extended Social MDPs we present are zero-shot, while most of
these prior approaches require social-interaction-specific training data.
We go well beyond such models supporting a far richer theory of social interactions.

Social actions such as walking, waving, hugging, and hand-shaking in videos of group activities have also been explored~\citep{patron2012structured,marin2014detecting,ryoo2009spatio}.
These methods broadly involve two phases~\citep{puig2020watch}: a social perception phase and a coordination phase where agents interact.
In contrast, Social MDPs are agnostic to specific social actions.
We are not detecting if two agents are hugging and then inferring that they are
friendly because we have seen that hugging often results in positive actions.
We are specifying a reward for what cooperative actions are; given a novel
action, that has never been seen before, in a novel context, Social MDPs
determine what if any social interaction is at play.

In game theory, several approaches \citep{levine1998modeling,feng2021bargaining} explore
altruistic and spiteful actions (the term of art for cooperation and conflict)
by means of linear combinations of pay-offs.
Work of \citet{levine1998modeling} is mathematically equivalent to the level 1 Social
MDPs defined in \citet{tejwani2021social}, while level 2 Social MDPs seem to
have no counterpart in the current economics literature.
This is the equivalent to saying that in a game you can help another player, but
you cannot help another player whose goal is to then help a third player.
Our long-term hope is to connect the extended Social MDPs presented here back to
the economics literature and expand both the depth of inference considered there
and the range of social interactions.

Interactive POMDPs \citep{doshi2009monte,doshi2008generalized,ng2012bayes}
(I-POMDPs) are the original blueprint for Social MDPs.
They extend POMDPs to allow agents to reason about other agent's beliefs.
They do not allow agents to reason about other agent's reward functions, making
it impossible for them to represent social interactions.
%
%

Human social perception has been studied extensively, but with limited
theoretical insights.
A rich history of work with animations of simple geometrical objects
\citep{heider1944experimental, abell2000triangles} shows that humans readily
assign social goals and intentions to any agent.
More recently, agents' behaviors have been simulated with modern physics engines
in fully observable \citep{shu2020adventures,kryven2021plans} and partially
observable environments \citep{netanyahu2021phase,puig2020watch} to create
various social benchmarks, although these benchmarks are almost exclusively
limited to helping and hindering.
We hope that by providing formal definitions for what social interactions are, we
can eventually bring about a much closer connection between the cognitive
science and robotics of social interactions.

\begin{figure*}
  \begin{minipage}{.60\textwidth}
  \scalebox{0.70}{%
    \begin{tikzpicture}
      \node[obs] (s_i) {$s$};
      \node[obs, below=2.2ex of s_i] (g_i) {$g_1$};
      \node[latent, below=2.2ex of g_i] (xi_i) {$\xi_1$};
      \node[latent, left=2.2ex of g_i] (a_i) {$\psi_1$};
      \plate [inner sep=.25cm,yshift=.2cm] {plate_i} {(g_i)(xi_i)(s_i)(a_i)} {};
      \edge {xi_i, g_i, s_i} {a_i};
      \node[left=9ex of xi_i] (yellow)
    {\includegraphics[width=.08\textwidth]{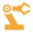}};
      \node[cloud callout,draw,inner sep=1pt,callout absolute pointer=(yellow.west)] (yellow-callout-red) at ($(yellow)+(-1.5, 0.7)$) {\includegraphics[width=.06\textwidth]{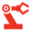}};
      \node[cloud callout,draw,inner sep=1pt,callout absolute pointer=(yellow-callout-red.east)] (yellow2) at ($(yellow)+(-0.2, 1.8)$) {\includegraphics[width=.05\textwidth]{images/yellow-robot.png}};
      \node[cloud callout,draw,inner sep=1pt,callout absolute pointer=(yellow2.west)] (yellow-tree) at ($(yellow-callout-red)+(-0.2, 1.7)$) {\includegraphics[width=.03\textwidth]{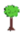}};

      \node[obs, right=18ex of xi_i, yshift=-10ex] (s_j) {$s$};
      \node[latent, below=2.2ex of s_j] (g_j) {$\widetilde{g}_2^{2,1}$};
      \node[latent, below=2.2ex of g_j] (xi_j) {$\widetilde{\xi}_2^{2,1}$};
      \node[latent, right=2.2ex of g_j] (a_j) {$\widetilde{\psi}_2^{1,1}$};
      \plate [inner sep=.25cm,yshift=.2cm] {plate_j} {(g_j)(xi_j)(s_j)(a_j)} {};
      \edge {xi_j, g_j, s_j} {a_j};
      \edge {g_j, xi_j} {xi_i};
      \node[right=10ex of xi_j] (red) {\includegraphics[width=.08\textwidth]{images/red-robot.png}};
      \node[cloud callout,draw,inner sep=1pt,callout absolute pointer=(red.east)] (red-callout-yellow) at ($(red)+(1.5, 0.7)$) {\includegraphics[width=.06\textwidth]{images/yellow-robot.png}};
      \node[cloud callout,draw,inner sep=1pt,callout absolute pointer=(red-callout-yellow.west)] (red-tree) at ($(red)+(0.3, 1.7)$) {\includegraphics[width=.035\textwidth]{images/tree.png}};

      \node[obs, left=18ex of xi_j, yshift=-10ex] (s_ij) {$s$};
      \node[latent, below=2.2ex of s_ij] (g_ij) {$\widetilde{g}_1^{1,2}$};
      \node[latent, left=2.2ex of g_ij] (a_ij) {$\widetilde{\psi}_1^{0,2}$};
      \plate [inner sep=.25cm,yshift=.2cm] {plate_ij} {(g_ij)(s_ij)(a_ij)} {};
      \edge {g_ij, s_ij} {a_ij};
      \edge {g_ij} {xi_j};
      \node[left=10ex of g_ij] (yellow3) {\includegraphics[width=.08\textwidth]{images/yellow-robot.png}};
      \node[cloud callout,draw,inner sep=1pt,callout absolute pointer=(yellow3.west)] (yellow-callout-tree) at ($(yellow3)+(-1.3, 0.7)$) {\includegraphics[width=.04\textwidth]{images/tree.png}};

      \node (level2) at ($(yellow)+(-3.5,1)$) {Level 2};
      \draw[dashed] (-6.5,-2.6)--(-4.5,-2.6);
      \node (level1) at ($(yellow)+(-3.5,-2.5)$) {Level 1};
      \draw[dashed] (-6.5,-6.2)--(-4.5,-6.2);
      \node (level0) at ($(yellow)+(-3.5,-5.7)$) {Level 0};

    \end{tikzpicture}
  }
  \caption{Reasoning with an extended Social MDP rendered as Bayesian network at
    different levels. $s$ are observations, $\psi$ is the policy, $g$ are
    physical goals, and $\xi$ are social goals. The yellow robot, agent 1, is
    reasoning at level two about the red robot's (agent 2) social actions. This
    demonstrates the utility of nesting Social MDPs, shallower social MDPs are
    more limited in their reasoning about the social abilities of other agents.}
  \label{fig:model-overview}
  \end{minipage}
  \begin{minipage}{.40\textwidth}
  \begin{algorithm}[H]
  \caption{The algorithm to compute social policy $\psi_i^l$ for agent $i$ at level $l$ and time $t$. We use the estimated social policy $\widetilde{\psi}_j^{l-1,i}$ at previous time step to update the estimated physical and social reward as described in \cref{sec:social-mdps}. At $t=0$, we assume $P(\widetilde{g}_{j}^{l,i,t})$ and $P(\widetilde{\xi}_{ji}^{l,i,t})$ are sampled from uniform distributions. This algorithm is called at all recursion steps $\widetilde{\psi}_j^{l-1,i}$ to estimate social policy for the other agent $j$. The estimated reward and policy are used to compute the Q values for selecting the actions.}
  \label{alg:algorithm}
  \begin{algorithmic}
  \Require $l, s^t, a_i^t, a_j^t, \xi_{ij}, g_i$
  \If{$l=0$}
    \State solve MDP for agent $i$
  \Else
    \ForAll{$\widetilde{\xi}_{ji}^{l,i,t}, \widetilde{g}_{j}^{l,i,t}$}
      \State \small{compute}
      \State \small{$P(\widetilde{\xi}_{ji}^{l,i,t}|s^{t-1}, a_i^{t-1}, a_j^{t-1})$}
      \State \small{$P(\widetilde{g}_{j}^{l,i,t}|s^{1:t-1})$}
      \State \small{$\widetilde{\psi}_j^{l-1,i}(s^t, a_j^t, a_i^t, \widetilde{\xi}_{ji}^{l,i,t}, \widetilde{g}_{j}^{l,i,t})$}
    \EndFor
    \State compute $R_i^l(s^t, a_i^t, a_j^t, \xi_{ij}, g_i)$
    \State compute $Q_i^l(s^t, a_i^t, a_j^t, \xi_{ij}, g_i)$
    \State $\pi_i^l \gets \argmax_{a_i \in \mathcal{A}_i} {Q_i^l}$
  \EndIf
  \end{algorithmic}
  \end{algorithm}
  \end{minipage}
\end{figure*}

\begin{table*}
  \centering
  \begin{tabular}{ccccc}
    Type of $\xi_{ij}^l$ & Type of $\widetilde{\xi}_{ji}^{l,i}$ & Substitute $\xi_{ij}^l$ at $l > 1$ & Substitute $\xi_{ij}^l$ at $l=1$  \\
    \midrule
    \midrule
    cooperation & any & $\widetilde{R}_j^{l-1,i}$ & $r(\widetilde{g}_j^{l,i})$ \\
    \midrule
    conflict & any & $-\widetilde{R}_j^{l-1,i}$ & $-r(\widetilde{g}_j^{l,i})$ \\
    \midrule
    \multirow{7}{*}{competition} & cooperation & $\widetilde{R}_j^{l-1,i}$ & \multirow{7}{*}{$-r(\widetilde{g}_j^{l,i})$} \\
    & conflict & $-\widetilde{R}_j^{l-1,i}$ & \\
    & competition & $-\widetilde{R}_j^{l-1,i}$ & \\
    \cmidrule{2-3}
    & coercion & $\begin{array}{cl}\widetilde{R}_j^{l-1,i} & \text{if } g_i=\widetilde{g}_j^{l,i} \\ -\widetilde{R}_j^{l-1,i} & \text{if } g_i \neq \widetilde{g}_j^{l,i}\end{array}$ & \\
    \cmidrule{2-3}
    & exchange & $\widetilde{R}_j^{l-1,i}$ & \\
    \midrule
    \multirow{7}{*}{coercion} & cooperation & $\widetilde{R}_j^{l-1,i} + r(g_i)$ & \multirow{7}{*}{$r(g_i)$} \\
    & conflict & $-\widetilde{R}_j^{l-1,i} + r(g_i)$ & \\
    \cmidrule{2-3}
    & competition & $\begin{array}{cl}\widetilde{R}_j^{l-1,i} & \text{if } g_i=\widetilde{g}_j^{l,i} \\ -\widetilde{R}_j^{l-1,i} + r(g_i) & \text{if } g_i \neq \widetilde{g}_j^{l,i}\end{array}$ & \\
    \cmidrule{2-3}
    & coercion & $-\widetilde{R}_j^{l-1,i} + r(g_i)$ & \\
    & exchange & $\widetilde{R}_j^{l-1,i} + r(g_i)$ & \\
    \midrule
    \multirow{2}{*}{exchange} & exchange & $n \cdot r(\widetilde{g}_j^{l,i}) + \widetilde{R}_j^{l-1,i}$ & \multirow{2}{*}{N/A} \\
    & others & N/A & \\
  \end{tabular}
  \caption{Computing the reward function of a social agent $i$, engaging with
    agent $j$ in one of five social interactions at level $l$. The reward
    function for $i$ is of the form $R^l_i = r(g_i) + \xi_{ij}^l$, where
    $r(g_i)$ is the reward for the physical goal of agent $i$ (the goal and
    reward in a plain MDP) and $\xi_{ij}^l$ is the social interaction of $i$ toward
    $j$. Based on the type of interaction that $i$ wants to have with $j$ and
    its estimate of what the reciprocal interaction is ($i$'s estimate of how
    $j$ wants to interact with $i$, $\widetilde{\xi}_{ij}^{l,i}$) we show what
    term to substitute $\xi_{ij}^l$ with achieve that interaction. For example,
    in cooperation, we substitute $i$'s estimate of what $j$ wants into $i$'s
    reward function; this gives $i$ a reward whenever $j$ accomplishes their
    goals. N/A are impossible interactions. See the text for an explanation of the other substitutions.}
  \label{tab:social-rewards}
\end{table*}

\section{Model}




We adopt the setting of \citet{tejwani2021social} who initially develop Social
MDPs by giving agents a combination of physical goals and social goals. A
physical goal is precisely what an MDP can represent, a reward that is a
function of the state of the world --- Social MDPs degenerate into
MDPs when the social goal is nil. A social goal extends the reward function
of every agent to allow it to contain an estimate of the reward function of
another agent. In the original Social MDP formulation, every agent estimated the
rewards of every other agent. Then, the reward of each agent was a linear
combination of its own physical goal and its estimate of the reward function of
other agents (the social goal).

The coefficient of this linear combination of physical and social goals
determined how willing the agent was to engage socially. If it was zero, no
weight was put on the social goal, so the agent behaved as an MDP. A large
positive value gives an agent a high reward when another agent maximizes its
own reward --- the result is an agent that helps the other agent succeed. A
large negative value does the opposite, gives an agent a low reward when another
agent maximizes its own reward --- the result is an agent that stops the other
agent from succeeding. \citet{tejwani2021social} demonstrated that this formulation gives
rise to robots that appear to humans to behave socially, and that this
coefficient determined the strength of that interaction and gave rise to rich
social behaviors.

Unfortunately, this framework lacks any degrees of freedom that would allow it
to represent any other social interactions aside from helping and hindering. It
has a single coefficient that determines the interaction between agents and the
polarity of that coefficient determines if the interaction is helpful or
unhelpful.

Next, we survey a precise mathematical definition of the five social interaction
types considered in this framework of recursively estimating other agent's
social reward functions and then demonstrate how to extend Social MDPs to enable
robots to execute these interactions.

\subsection{The five interaction types}

To define the pairwise interactions between agent $i$ and $j$ let the reward of
an agent $i$ at level $l$, $R^l_i$, be a combination of that agent's physical
goal, $g_i$ and that agent's social goal, $\xi_{ij}^l$;
$R^l_i = r(g_i) + \xi_{ij}^l$. Let $\xi_{ij}^l$ be the social goal of agent $i$
toward agent $j$ at level $l$. At $l=0$, no agents are social, the Social MDP
degenerates into a traditional MDP, i.e., $\xi_{ij}^0=0$. At $l=1$, each agent
has a social goal, but it does not consider other agents to have social goals,
only physical goals. At $l=2$, agents assume that other agents also have social
goals, etc. This nesting allows Social MDPs to start with an arbitrary level,
then eventually bottom out in an MDP. Next, we will show how the reward function
of agent $i$ is formulated when that agent engages in each of the five social
interaction types.

We define all interactions from the perspective of agent $i$, i.e., if agent $i$
wants to be social with respect to agent $j$ in each of these five ways, what is
$i$'s reward function? Reward functions are specified in terms of estimated
rewards, as $i$ does not know what $j$ wants, it must infer $j$'s reward
function; we denote all estimated quantities with a tilde, as in
$\widetilde{R}^{l-1,i}_j$ for the estimate of $j$'s reward function made by $i$
at level $l-1$ and $\widetilde{\xi}_{ij}^{l,i}$ for the estimate of the social
interaction between $i$ and $j$ at level $l$ again made by $i$. In each case, the
additional superscript $i$ denotes the agent performing the estimation ---
agents $i$ and $k$ may, depending on their observations or biases, not estimate
the same reward function for agent $k$.

An overview of the recursive inferences being made between two agents is shown
in \cref{fig:model-overview}. In \cref{tab:social-rewards}, we show the social
component of each of the reward functions of agent $i$ as a function of its own
social goal and the estimated social goal of $j$ when $l>1$ (when other agents
are assumed to have social goals) and when $l=0$ (when other agents have only
physical goals). For each agent, we estimate what kind of social interaction it
is engaging in. Then, the reward function of the agent is updated according to
\cref{fig:model-overview}. For example, if level 2 agent $i$ intends to
cooperate with $j$, the original reward function template for
$i$, $R^l_i = r(g_i) + \xi_{ij}^l$, is updated as
$R^l_i = r(g_i) + \widetilde{R}_j^{l-1,i}$ --- agent $i$'s reward now includes
agent $j$'s reward so $i$ will help $j$.

\emph{Cooperation} and \emph{Conflict}, the two types of goals supported by the
original Social MDPs are peculiar: the reward function does not depend on what
kind of social interaction the other agent wants to engage in. This is why they
could be implemented in the original Social MDP formulation. The other three
types of social interaction are more complex, they depend on the other agent's
intentions. For example, competing with someone who intends to
cooperate with you is very different from competing with someone who intends to
stop you from taking actions.

%

See \cref{tab:social-rewards} for the complete mapping from interactions to rewards.
\emph{Competition} and \emph{Coercion} follow a pattern that is similar to cooperation and conflict with several special cases since they are situation-specific.
In other words, what it means to compete depends on what the other agent wants to do, if they are pleasant you incorporate their reward differently into yours than if they are combative.

\emph{Exchange} is the outlier.
Exchanging implies that an agent performs an action that is useful for another and vice versa.
If one of the two agents isn't willing to exchange, it's impossible for either to do so on their own hence the N/A in \cref{tab:social-rewards}.
When directing an agent to exchange with another, we update that agent's reward function to include a copy of the other agent's goal with a positive, but small, scalar $n$.
This makes the agent willing to help, weakly, while maintaining its own goals.
If both agents do this symmetrically, the result is a pair which is willing to perform mutually-beneficial actions as long as they can accomplish their other goals more effectively.

In the results section we demonstrate that the method outlined in \cref{tab:social-rewards} of mapping different social interactions to reward functions gives rise to behavior which is interpreted as the correct social interaction by humans. Next we describe how to build and perform inference with an extended Social MDP, one which can handle this richer set of social interactions and which estimates the type of interaction rather than merely supporting cooperation and conflict.

%

%
%
%
%
%

\subsection{Extended Social MDPs}

A Social MDP for agent $i$ with respect to all agents $J$ consists of an arity
(here we formulate the pairwise case) and a maximum level, $l$, and is defined
as:
\begin{equation}
    \label{eq:social-mdp}
    M_{i}^{l} = \langle \mathcal{S}, \mathcal{A}, T, \xi_{ij}, g_i, R^l_i, \gamma \rangle
\end{equation}
where
\begin{itemize*}[label={}, itemjoin={;}, itemjoin*={; and}]
    \item $\mathcal{S}$ is a set of states in the environment where $s \in \mathcal{S}$
    \item $\mathcal{A} = \mathcal{A}_i \times \mathcal{A}_j$ is the set of joint moves of agent $i$ and agent $j$
    \item $T$ is the probability distribution of going from state
        $s \in \mathcal{S}$ to next state $s^{\prime} \in \mathcal{S}$ given
        actions of both agents: $T\left(s^{\prime} \mid s, a_i, a_j \right)$
    \item $\xi_{ij}$ is agent $i$'s intended social interaction with agent $j$. It computes agent $i$'s social reward when interacting with agent $j$
    \item $g_i$ is agent $i$'s physical goal
    \item $R^l_i$ is the $l$-th level reward function for agent $i$ based on its estimate of other agents' rewards
    \item $\gamma$ is a discount factor, $\gamma \in (0,1)$.
\end{itemize*}

Each agent has its own physical goal, e.g., going to a landmark, as well a social goal,
e.g., helping or hindering other agents.
What enables Social MDPs to go beyond regular MDPs is the recursive nature of the reward function
which can be written in terms of the estimated rewards of other agents.
The immediate reward of an agent $i$ at each time step is computed as follows:
\begin{equation} \label{eq:reward}
  R^l_i(s, a_i, a_j, \xi_i, g_i) = r(g_i) + \xi_i(g_i, \widetilde{g_j}, \widetilde{\xi_j})
  - c(a_i)
\end{equation}

\subsection{Planning with Social MDPs}
\label{sec:social-mdps}

Analogous to MDPs, the Q function of Social MDPs is the sum of immediate reward and the expected value in the future.

\begin{equation}
\begin{split}
    Q_i^l(s, a_J, \xi_{ij}, g_i) & =
    R(s, a_i, \xi_{ij}, g_i)\\
    & + \gamma \sum_{s^{\prime} \in S} T(s, a_J, s^{\prime}) V_i^l(s', \xi{ij}, g_i)
\end{split}
  \end{equation}
Since agent $i$ is interacting with other agents $j \in J$, it needs to
estimate what actions other agents are likely to take in order to compute its state-action value.
Social MDPs take the expectation over the estimated goals and actions of
agent $j$ to compute $V_i^l(s', \xi{ij}, g_i)$:

\begin{equation}
\scalebox{0.8}{$\begin{split}
\label{eq:v-func}
  V_{i}^{l}(s^{\prime}, \xi_{ij}, g_i)
  {}&=
  \max_{a_i^{\prime} \in \mathcal{A}_i} \Bigg\{
    E_{\widetilde{g}_j^{l,i}, \widetilde{\xi}_{ji}^{l,i}, a_j^{\prime}}
    [Q_i^l(s^{\prime}, a_J^{\prime}, \xi_{ij}, g_i)] \Bigg\}\\
  {}&=
  \max_{a_i^{\prime} \in \mathcal{A}_i} \Bigg\{
    \sum_{\substack{j \in J,\\j \neq i}} \sum_{a_j^{\prime} \in \mathcal{A}_j}
    \sum_{\widetilde{g}_j^{l,i}} \int_{\widetilde{\xi}_{ji}^{l,i}}
    \underbrace{P(\widetilde{g}_j^{l,i} | s^{1:t})}_{\substack{\text{estimate physical goal}\\ \text{(Eq.~\ref{eq:goal-est})}}} \\&
    \underbrace{P(\tilde{\xi}^{l,i}_{ji} \mid s, a_J)}_{\substack{\text{estimate social goal}\\ \text{(Eq.~\ref{eq:chi_est})}}}
    \underbrace{\widetilde{\psi}_{j}^{l-1,i}(s^{\prime}, a_J^{\prime}, \widetilde{\xi}^{l,i}_{ji}, \widetilde{g}_j^{l,i})}_
    {\substack{\text{estimate social policy}\\ \text{(Eq.~\ref{eq:si-policy})}}}
    Q_i^l(\cdot)
    d\widetilde{\xi}^{l,i}_{ji} \Bigg\}\\
\end{split}$}
\end{equation}

When solving agent $i$'s MDP at level $l$, the estimated social and physical goals are further used to update the other agent $j$'s social policy to the actions agent $j$ may take.
We denote the estimated social policy for agent $j$ at reasoning level $l-1$ as $\widetilde{\psi}_{j}^{l-1,i}: \mathcal{S} \times \mathcal{A}_J \times \widetilde{\xi}^{l,i}_{ji} \times \widetilde{g}_j^{l,i}  \rightarrow [0,1]$.
\cref{alg:algorithm} summarizes the steps to compute the state-action values and select optimal actions for any level $l$ at time step $t$.
We first update the probability of the estimated goals of other agents using the observed state and the estimated policy from the previous time step.
The updated probability of goals are used to update the policy of other agents and compute the reward and Q function of the target agent.
%

An agent's estimate of another agent's social and physical goals at time step $t$ and level $l$ can be updated based on the actions performed by the agents.
At $t=0$, we use uniform distributions for social and physical goals.
The social goal, estimated at time step $t$, is updated after actions
taken by all agents at the previous time step.
This update is similar to the belief update in the POMDP framework but
based on the estimated social policy of the other agent $j$:

\begin{equation}
\label{eq:chi_est}
\scalebox{0.9}{$\begin{split}
    P(\widetilde{\xi}_{ji}^{l,i,t} \mid s^{t-1}, a_J^{t-1}) \propto
    P(\widetilde{\xi}_{ji}^{i,{t-1}} \mid s^{t-2}, a_J^{t-2}) \\
    \sum_{\widetilde{g}_j^{i,l,t-1}}
    P(a_j^{t-1} \mid s^{t-1}, \widetilde{\xi}_{ji}^{i,l,{t-1}}, \widetilde{g}_j^{i,l,t-1})
    \times T(s^{t-1}, a_J^{t-1}, s^{t})
\end{split}$}
\end{equation}

The physical goal $g_j$ of agent $j$ is estimated by $i$ as follows, similar to \citep{shu2020adventures} but marginalized over the estimated social goal as the agent is estimating the social goal at the same time.
\begin{equation} \label{eq:goal-est}
\begin{split}
    P(\widetilde{g}_j^{l,i,t} | s^{1:t-1}) \propto
    \int_{\widetilde{\xi}_{ji}^{l,i,t}}
    P(s^{1:t-1}|\widetilde{g}_j^{l,i,t}, \widetilde{\xi}_{ji}^{l,i,t})\\
    P(\widetilde{g}_j^{l,i,t})P(\widetilde{\xi}_{ji}^{l,i,t})~d\widetilde{\xi}_{ji}^{l,i,t}
\end{split}
\end{equation}

The $l$-level social policy $\widetilde{\psi}_{j}^{l,i}$ of the
agent $j$ is predicted by $i$ using the Q-function at level $l$-1:
\begin{equation}
\label{eq:si-policy}
    \widetilde{\psi}_{j}^{l-1,i}(s, a_J, \widetilde{\xi}_{ji}^{l,i}, \widetilde{g}_j^{l,i}) =
    \text{Softmax}(Q_j^{l-1}(s, a_J, \widetilde{\xi}_{ji}^{l,i}, \widetilde{g}_j^{l,i}))
\end{equation}

This is a softmax policy where we use a temperature parameter $\tau$ to
control how much the agent $j$ follows greedy actions.
As shown in  eq. \ref{eq:v-func}, in order to use agent $j$'s Q
function at level $l$-1, it requires to compute agent $i$'s Q
function at level $l$-2, and so on.
Recursively solving Social MDPs eventually bottoms out in level 0 where one
solves an MDP.

\begin{table*}[t]
\centering
\begin{tabular}{lccccccc}
  \text { Model } & \multicolumn{6}{c} {\text { Social Goal }} & \text { Physical Goal } \\
  \cmidrule{2-7}
                  & \text { Cooperation } & \text { Conflict } & \text { Competition } & \text { Coercion } & \text { Exchange } & \text { \textbf{Overall} } &\\[1ex]
  \text { \emph{Human} } & \emph{0.934} & \emph{0.952} & \emph{0.623} & \emph{0.724} & \emph{0.876} & \emph{0.823} & \emph{0.974}\\
  \text { Extended Social MDP (Ours) } & \textbf{0.845} & \textbf{0.851} & \textbf{0.471} & \textbf{0.651} & \textbf{0.814} & \textbf{0.726} & \textbf{0.831} \\
  \text { Inverse Planning} & 0.763 & 0.784 & 0.261 & 0.283 & 0.197 & 0.457 & 0.783\\
  \text { Cue Based Model} & 0.461 & 0.434 & 0.127 & 0.156 & 0.083 & 0.252 & 0.432\\
\end{tabular}
\caption{The accuracy of humans and each of the models at determining which social interaction is taking place in each of the 72 scenarios. Our model is significantly more accurate, particularly when it comes to recognizing social interactions.}
\label{fig:model-human-accuracy}
\end{table*}

\begin{figure*}[t]
    \centering
    \vspace{-3ex}
    \includegraphics[width=0.77\textwidth]{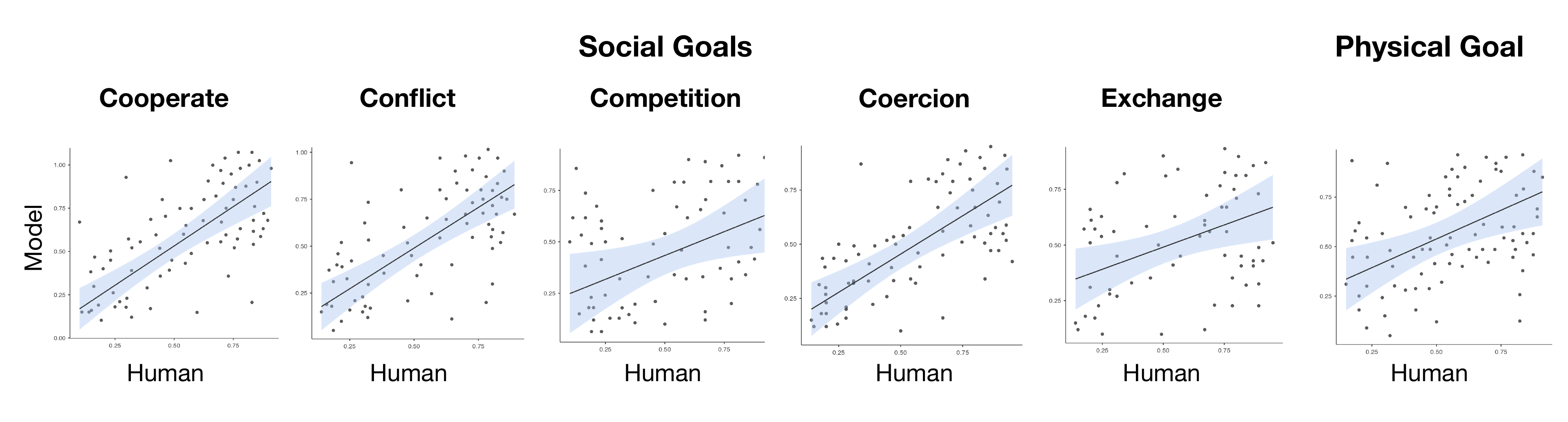}\\[-2ex]
    \caption{Humans and our model scored 72 scenarios according to how
      likely each social interaction was and how likely one of the physical
      goals was. The straight line is the best linear fit and the light blue band
      represents the 95\% confidence interval. Our model agrees with
      humans and predicts their confidence scores for both social
      interactions and the physical goal.}
\label{fig:social-goals-weights}
\end{figure*}

\section{Results}

We apply our extended Social MDP framework to a multi-agent gridworld inspired
by previous studies on social perception
\citep{ullman2009help,baker2008theory,baker2011bayesian,netanyahu2021phase}.
This $10 \times 10$ world consists of two agents (a yellow robot and red robot),
two physical landmarks (a construction site and a tree) and three objects (an
axe, wooden log, and a water bucket).
Objects can be pushed to either destination.
Physical goals consist of moving the desired objects to one of the landmarks.
Agents can have no social goal or one of the five social goals: cooperation,
conflict, competition, coercion, or exchange.
At any point in time, agents can push an object forward, move in one of the four
cardinal directions or choose to take no actions.

Each agent's
reward for reaching its physical goal is based on that agent's geodesic distance
from the goal after taking an action \cite{ullman2009help}.
This physical reward function is parameterized by $\rho$ and $\delta$ that
determines the scale and shape of the physical reward:
$r(s,a,g_i)=\max \left(\rho\left(1-\operatorname{distance}(s, a, g_i) / \delta\right), 0\right)$.
We set the cost, $c$, of an action $a$, to 1 for grid moves and to 0.1 for staying in place while $\rho$ and $\delta$
were set to 1.25 and 5, respectively.
The discount factor, $\gamma$, was set to $0.99$.

We systematically enumerate the 72 unique scenarios by assigning agents either
the same or different physical goal, and one of five social goals or no social
goal ($2*6*6=72$ scenarios).
Even this simple world gives rise to many rich social interactions with subtle
but meaningful differences in the behaviors of agents depending on their
understanding of social interactions.
For instance, take the scenario where both agents have the social goal of
exchanging something.
An exchange is practical only in some cases: the agents must be arranged in such
a way that it would be helpful for the red agent to aid the yellow agent and
vice versa (for example, the red agent is closer to the axe which the yellow
agent wants and the yellow agent is closer to the log which the red agent
wants).
Agents must recognize this.
Then they must recognize if the other agent is willing to exchange by attempting
the exchange.
Then, they must follow through with the exchange, bring the other object to a
meeting point, swap objects, and then go on to complete their own physical goals.
If one of the agents is uncooperative, both agents should abort and try to
perform their own physical goals, as the exchange is then impossible.
This occurs without pre-specifying any symbolic notion of ``exchange'', without
a single training example of an exchange, and without any hardcoded rules,
merely by specifying the correct reward function.

All scenarios with descriptions, diagrams, videos, and detailed
timestep-by-timestep results for all human experiments and models are available
at this URL

{\blue\scriptsize\url{https://social-interactions-mdp.github.io}}
%

\begin{figure*}[t]
    \centering
    \vspace{0ex}
    \includegraphics[width=0.8\textwidth]{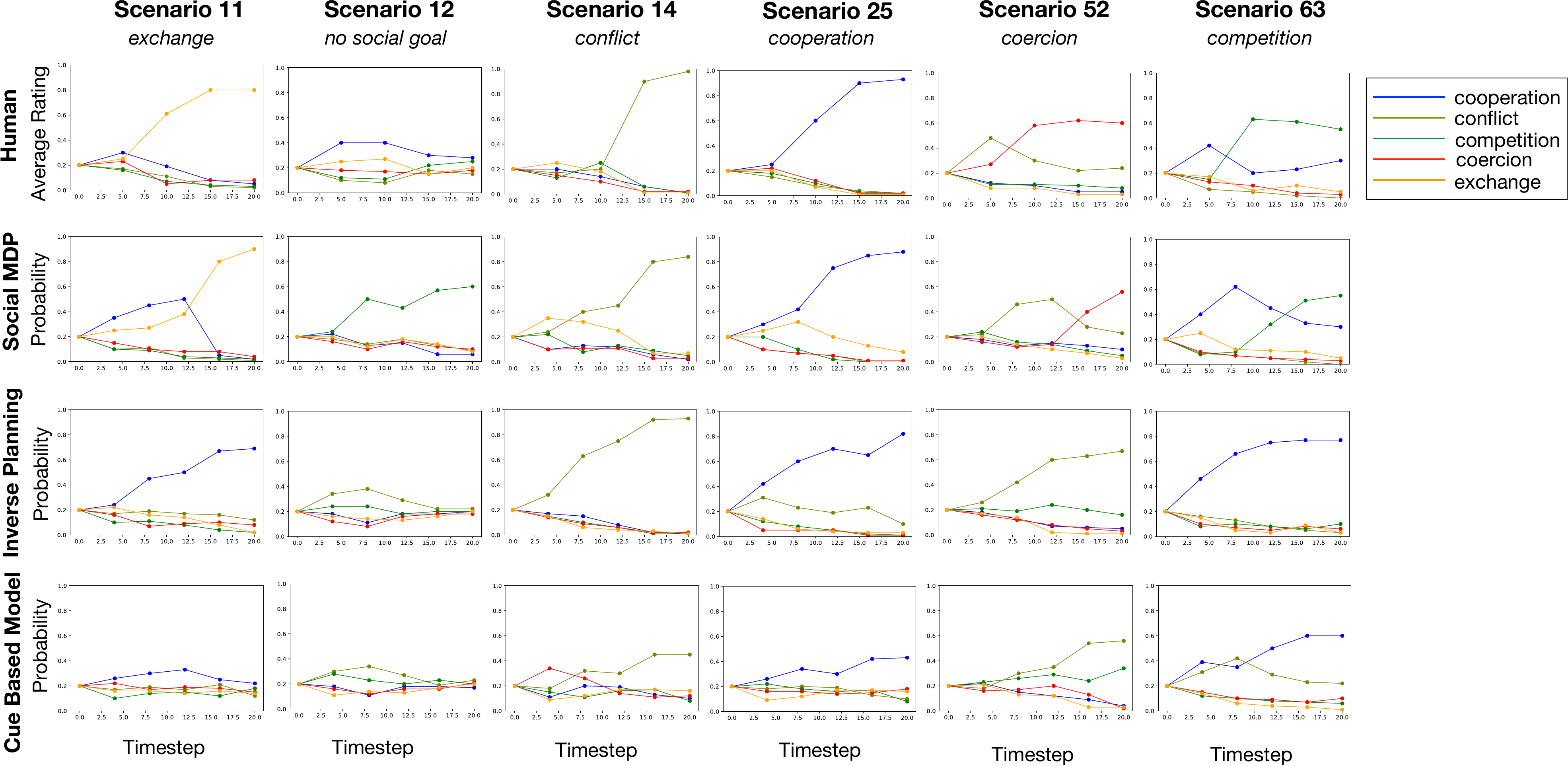}\\
    \label{fig:social-interactions-estimates}
    \caption{Humans, Social MDPs, and the two other baseline models were asked
      predicted the social interaction in each scenario at every time step. We
      chose six representative scenarios that each show a different interaction.
      Above every column we provide the scenario number along with its intended
      social interaction (our website provides rendering of every scenario along
      with many other details for each).
      Social MDPs agree with humans timestep by timestep, with the exception of
      scenario 12 where humans recognize that there is no social interaction
      while Social MDPs are $\approx$60\% certain that agents are competing.}
      \vspace{-2ex}
\end{figure*}

\subsection{Are these social interactions?}

We first establish that these scenarios do indeed show social interactions and
that humans are able to recognize the intended social interactions.
Each of the 72 scenarios were executed with level 2 Social MDPs guiding the
yellow agent giving rise to 72 videos.
12 subjects were recruited on Mechanical Turk.
Each subject saw each of the 72 videos and rated it according to their
confidence that the video depicts a given social interaction after each action
taken by every agent.
Taking the human subject's last judgment, at which point they have seen the
entire interaction, we ask how well does the most confident label that humans
give to the interaction match the intended interaction from the Social MDP.
Humans provided the top label with 82\% accuracy, showing that Social MDPs
generate the intended social interactions the vast majority of the time, see
\cref{fig:model-human-accuracy}.

\subsection{Do we model human judgments?}

Going further, we demonstrate that not only do humans recognize the intended
social interactions from Social MDPs, but Social MDPs also predict human
judgments.
Taking each of the 72 videos, we compare the confidence of humans that a
particular social interaction is being depicted to that of the model.
We find that our model has a 0.784 correlation with human judgments (0.86 for
cooperation, 0.84 for conflict, 0.64 for competition, 0.64 for coercion, 0.76
for exchange), see \cref{fig:social-goals-weights} in blue.
When considering estimating the physical goal, rather than the social goal, our
model has a 0.76 correlation with human judgments.

\subsection{Models}

We compare Social MDPs with two alternative models, inverse planning
\citep{ullman2009help} and a recent cue-based model \citep{shu2020adventures}.
Each model is provided with every video, frame by frame, and is required to
incrementally predict which social interaction is taking place and what the
physical goal is.
At our website, we list every scenario with detailed results from each model at
every time step.
Several example scenarios are shown in \cref{fig:social-interactions-estimates}.
Qualitatively, one can observe that the Social MDPs track human judgments about
these videos far more accurately.
Quantitatively, results aggregated across the different categories are shown in
\cref{fig:model-human-accuracy}.
While humans are able to estimate social goals and physical goals more
accurately than any model, the gap between Social MDPs and other models is stark.
Not only do Social MDPs estimate the physical goal more accurately, but the
social goal accuracy is dramatically higher (72.6\% vs 45.7\%).
Our online supplement also demonstrates the different levels of nesting.

\section{Conclusion}

We demonstrated extended Social MDPs which are able to reason about the five
core social interactions described in the microsociology literature.
This model is capable of reasoning about social interactions without requiring
interaction-specific training, just like humans are able to recognize social
interactions in novel environments.
The inferences the model makes both qualitatively and quantitatively match human
judgments.
This significantly expands the space of social interactions that robots engage in.
The runtime of the model depends on the number of levels considered as this
controls how many MDPs must be solved recursively.
We do not yet know which level corresponds to the inferences that humans are
able to make.

We would like to scale this approach up to continuous environments and to
physical robots with manipulators.
In addition, we would like to recognize these social interactions from
real-world videos rather than from renderings and trajectories of agents.
Hopefully, sometime in the future, robots will have general-purpose social
skills and the field of studying social interactions will be based on
mathematical and refutable theories that span robotics, social science,
cognitive science, vision, and neuroscience.

\clearpage
\section*{Acknowledgments}
This work was supported by the Center for Brains, Minds and Machines, NSF STC award 1231216, the MIT CSAIL Systems that Learn Initiative, the CBMM-Siemens Graduate Fellowship, the MIT-IBM Watson AI Lab, the DARPA Artificial Social Intelligence for Successful Teams (ASIST) program, the United States Air Force Research Laboratory and United States Air Force Artificial Intelligence Accelerator under Cooperative Agreement Number FA8750-19-2-1000, and the Office of Naval Research under Award Number N00014-20-1-2589 and Award Number N00014- 20-1-2643. The views and conclusions contained in this document are those of the authors and should not be interpreted as representing the official policies, either expressed or implied, of the U.S. Government. The U.S. Government is authorized to reproduce and distribute reprints for Government purposes notwithstanding any copyright notation herein.

\bibliographystyle{IEEEtranN}
\renewcommand*{\bibfont}{\footnotesize}
\bibliography{references}  

\clearpage


\end{document}